\documentclass[10pt,twocolumn,letterpaper]{article}

\usepackage{cvpr}
\usepackage{times}
\usepackage{epsfig}
\usepackage{graphicx}
\usepackage{amsmath}
\usepackage[caption=false,font=footnotesize]{subfig}
\usepackage{amssymb}
\usepackage{tabularx}
\usepackage{booktabs}
\usepackage{multirow}
\usepackage{multicol}
\usepackage{csquotes}
\usepackage{rotating}
\usepackage{caption}
\usepackage{float}
\usepackage{array, makecell}
\usepackage[flushleft]{threeparttable}
\usepackage{tablefootnote}
\usepackage{booktabs}
\usepackage[export]{adjustbox}
\usepackage{array}
\newcolumntype{P}[1]{>{\centering\arraybackslash}p{#1}}
\newcolumntype{M}[1]{>{\centering\arraybackslash}m{#1}}
\usepackage{dblfloatfix}
\usepackage[pagebackref=true,breaklinks=true,letterpaper=true,colorlinks,bookmarks=false]{hyperref}

\hypersetup{
  colorlinks, linkcolor=red
}
\makeatletter
\g@addto@macro{\UrlBreaks}{\UrlOrds}
\makeatother

\usepackage{capt-of,etoolbox}

\DeclareRobustCommand{\etal}{\textit{et al.}\@\xspace}
\renewcommand{\mkbegdispquote}[2]{\itshape}

\cvprfinalcopy 

\def\cvprPaperID{} 

\ifcvprfinal\fi
\begin{document}

\title{Infant-Prints: Fingerprints for Reducing Infant Mortality}

\author{Joshua J. Engelsma, Debayan Deb, Anil K. Jain\\
Michigan State University\\
East Lansing, MI, USA\\
{\tt\small \{engelsm7, debdebay, jain\}@cse.msu.edu}
\and
Prem S. Sudhish, Anjoo Bhatnager\\
DEI, Saran Ashram Hospital\\
Agra UP 282005, India\\
{\tt\small pss@alumni.stanford.edu, dranjoo@gmail.com}
}

\twocolumn[{%
\renewcommand\twocolumn[1][]{#1}%
\maketitle
\thispagestyle{empty}
\begin{center}
    \centering
     \begin{minipage}{0.3\linewidth}
    \includegraphics[width=\linewidth]{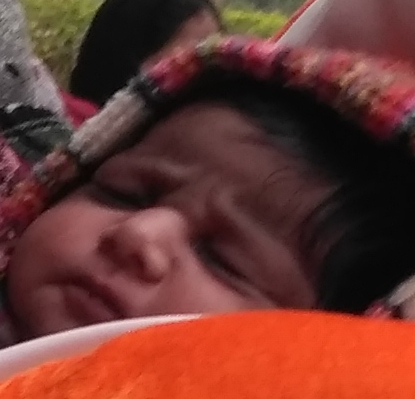}
    \centering
    \end{minipage}\;\;\;\;
    \begin{minipage}{0.28\linewidth}
    \includegraphics[width=\linewidth]{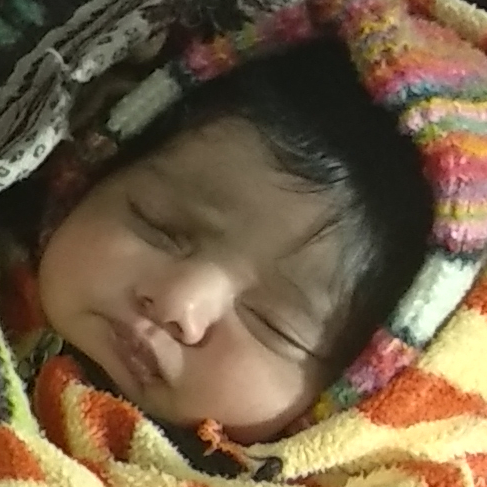}
    \centering
    \end{minipage}\;\;\;\;
    \begin{minipage}{0.28\linewidth}
    \includegraphics[width=\linewidth]{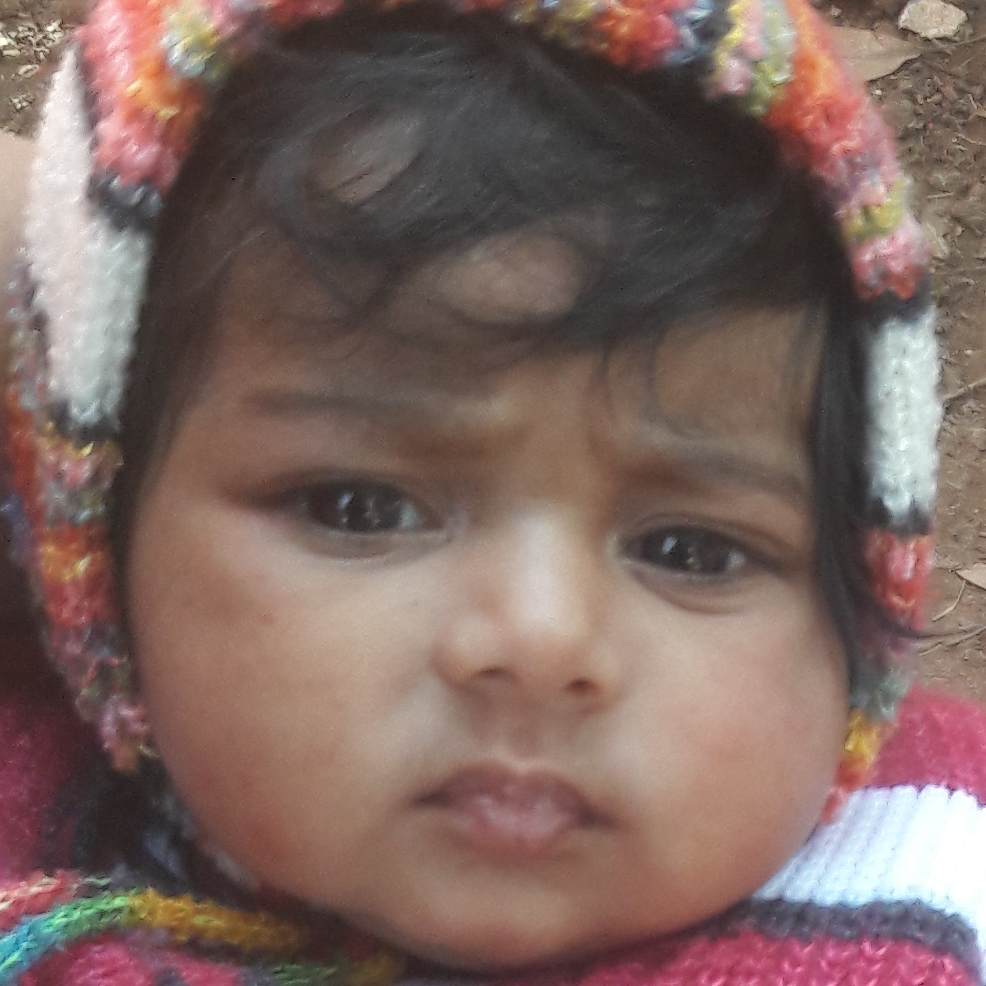}
    \centering
    \end{minipage}\\
    \begin{minipage}{0.28\linewidth}
    \includegraphics[width=\linewidth]{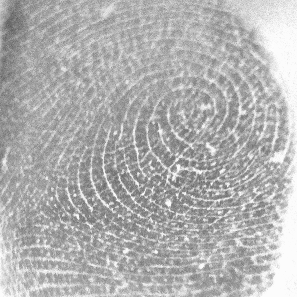}
    \centering {\\(a) 13 days old}
    \end{minipage}\;\;\;\;
    \begin{minipage}{0.28\linewidth}
    \includegraphics[width=\linewidth]{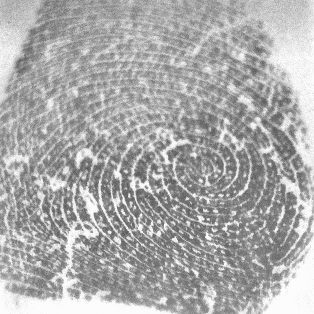}
    \centering {\\(b) 15 days old}
    \end{minipage}\;\;\;\;
    \begin{minipage}{0.28\linewidth}
    \includegraphics[width=\linewidth]{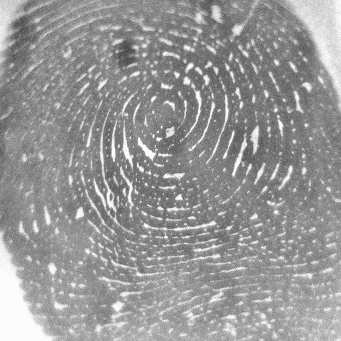}
    \centering {(c) 3 months and 5 days old}
    \end{minipage}\\
    \begin{minipage}{\linewidth}
    \begin{figure}[H]
    \captionof{figure}{\emph{Have we seen this infant before? Is this the child who her parents claim her to be?} Face images and corresponding left thumb fingerprints of an infant, \emph{Maanvi Sharma}, acquired on (a) December 16, 2018 (13 days old), (b) December 18, 2018 (15 days old), and  (c) March 5, 2019 (3 months and 5 days old) at Saran Ashram Hospital, Dayalbagh, India.}
    \label{fig:frontpage}
    \end{figure}
    \end{minipage}
\end{center}%
}]

\begin{abstract}
\vspace{-1.0em}
In developing countries around the world, a multitude of infants continue to suffer and die from vaccine-preventable diseases, and malnutrition. Lamentably, the lack of any official identification documentation makes it exceedingly difficult to prevent these infant deaths. To solve this global crisis, we propose Infant-Prints which is comprised of (i) a custom, compact, low-cost (85 USD), high-resolution (1,900 ppi) fingerprint reader, (ii) a high-resolution fingerprint matcher, and (iii) a mobile application for search and verification for the infant fingerprint. Using Infant-Prints, we have collected a longitudinal database of infant fingerprints and demonstrate its ability to perform accurate and reliable recognition of infants enrolled at the ages 0-3 months, in time for effective delivery of critical vaccinations and nutritional supplements (TAR=90\% @ FAR = 0.1\% for infants older than 8 weeks).
\end{abstract}


\section{Introduction}
It is estimated that there are more than 600 million children living worldwide between the ages of 0-5 years~\cite{age_structure}. Everyday, over 353 thousand newborns set foot on the planet~\cite{birth_rate}, with a majority of these births taking place in the poorest regions of the world. It is likely that neither the infants nor their parents will have access to any official identification documents, and consequently, efficient delivery and fraud prevention of healthcare, immunization, and nutrition supply are incredibly challenging. This is especially problematic for infants\footnote{Infants are considered to be in the 0-12 months age range, whereas, toddlers and preschoolers are within 1-3 and 3-5 years of age, respectively~\cite{age_classification}.} (0-1 years of age), when the child is at their most critical stage of development. 

Even with a growing world population, global vaccination coverage has remained constant in recent years. For instance, from 2015 to 2018, the percentage of children who have received their full course of three-dose diphtheria-tetanus-pertussis (DTP3) routine immunizations remains at 85\% with no significant changes~\cite{global_immunization}. This falls short of the Global Vaccine Action Plan's (GVAP\footnote{\url{https://bit.ly/1i7s8s2}}) target of achieving global immunization coverage of 90\% by 2020. According to the World Health Organization (WHO), inadequate monitoring and supervision, and lack of official identification documents (making it difficult to accurately track vaccination schedules) are key factors\footnote{\url{https://bit.ly/1pWn6Gn}}. 

Infant recognition is also necessary to effectively provide nutritional supplements. For example, the World Food Programme (WFP) found that in Yemen, a country with 12 million starving residents, food distribution records are falsified and relief is being given to people not entitled to it, preventing those who actually need aid from receiving it~\cite{fraud2},~\cite{fraud3}. Finally, infant fingerprint recognition would aid in baby swapping prevention\footnote{\url{https://bit.ly/2U5eAHn}}, identifying missing children, and access to government benefits, healthcare, and financial services throughout the infant's lifetime. 

\begin{figure}[!t]
    \centering
    \includegraphics[width=\linewidth]{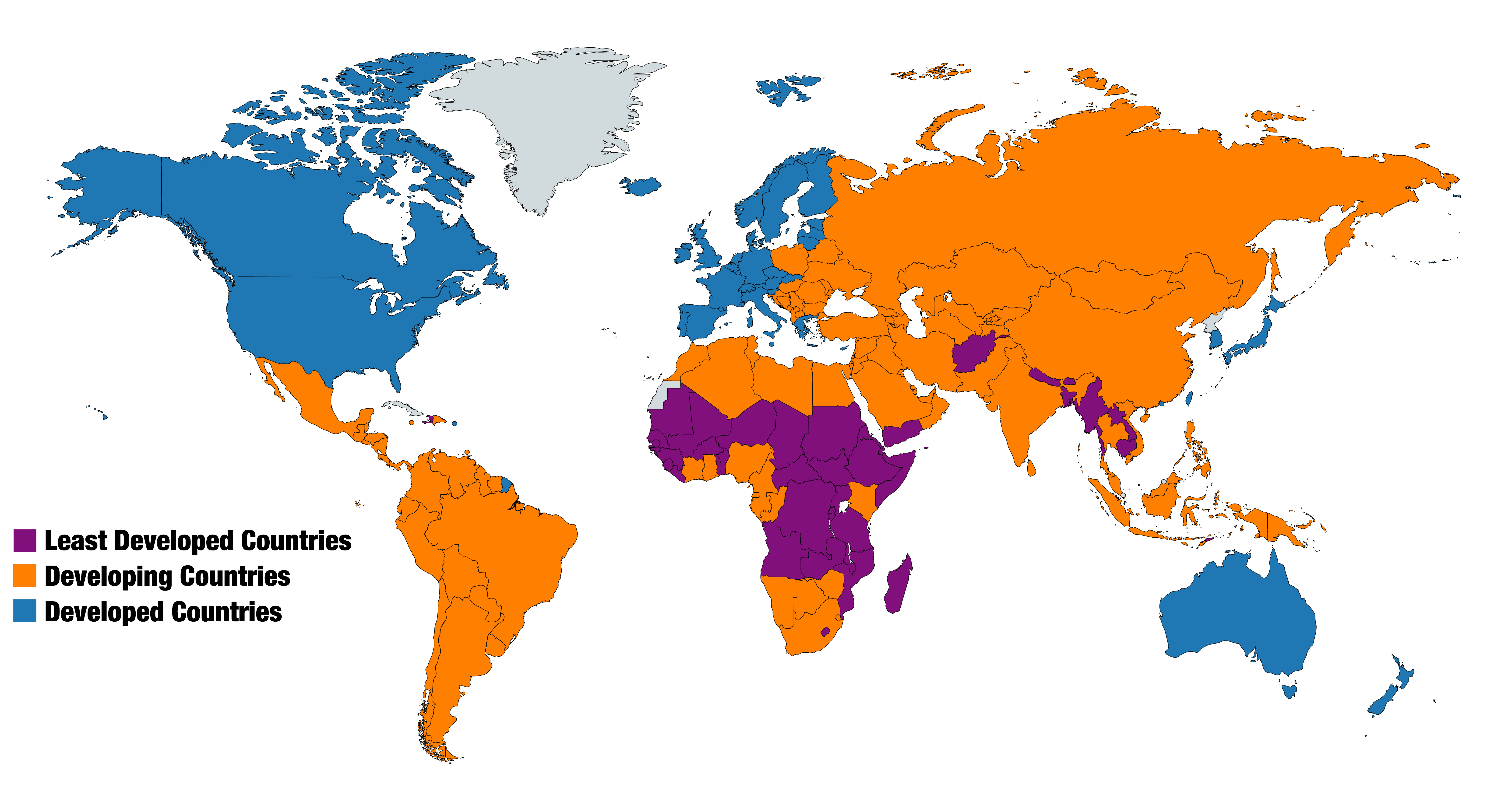}
    \caption{The countries highlighted in purple, orange, and blue denote the least developed (LDC), developing, and developed countries, respectively, according to the United Nations~\cite{ldr}. Classification is done according to poverty levels (Gross National Income per capita $<$ USD 1,025 for LDC), human resource weakness (nutrition, health, education and literacy), and economic vulnerability. As of February 2019, there are 47 least developed, 92 developing, and 54 developed countries in the world~\cite{country_un},~\cite{country_imf}.}
    \label{fig:ldr}
\end{figure}

Conventional identification documents (paper records) are impractical because they may be fraudulent~\cite{fraud} or become lost or stolen. This motivated India's ambitious and highly successful national ID program, called \emph{Aadhaar}, which uses biometric recognition (a pair of irides, all ten fingerprints and face) to uniquely identify (de-duplicate) and then authenticate over 1.2 billion Indian residents\footnote{\url{https://bit.ly/2zqrBSq}} that are over the age of 5 years. However, due to a lack of accurate and reliable biometric recognition of infants, the youngest among us still remain incredibly vulnerable, especially those living in least developed and developing\footnote{The United Nations classifies countries into three broad categories: (i) Least Developed, (ii) Developing, and (iii) Developed~\cite{ldr}.} countries (Fig.~\ref{fig:ldr}). Notably, 36\% of the population in low-income economies lack official IDs, compared to 22\% and 9\% in lower-middle and upper-middle income economies~\cite{global_identity} and 17\% of those lacking identification are under the age of five~\cite{country_un}.

Designing a biometric recognition system for infants is a significant challenge in part due to the fact that a majority of the biometric modalities are not useful for infants. An infant's face changes daily as they grow. Iris image capture is also not feasible for infants (child is sleeping or crying). Footprint recognition~\cite{footprint}, requires removing socks and shoes and cleaning the infant's feet. 

\begin{figure}[!t]
    \centering
    \subfloat[]{\includegraphics[height=1.05in]{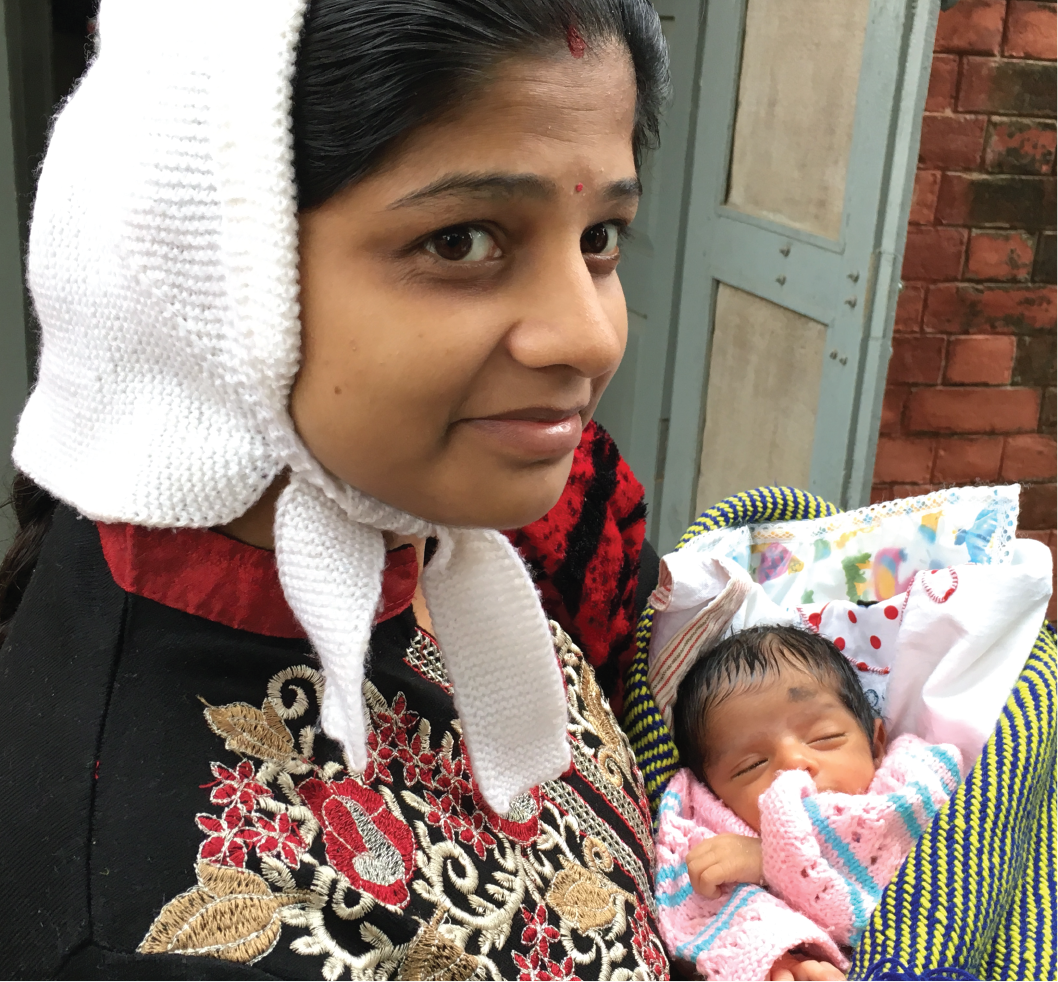}}\hfill
    \subfloat[]{\includegraphics[height=1.05in]{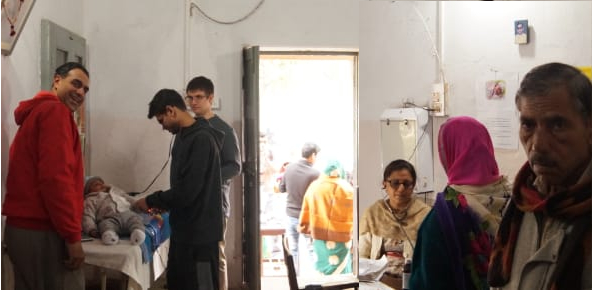}\label{fig:bhatnagar}}
    \caption{Infant fingerprint collection at Saran Ashram hospital, Dayalbagh, India. (a) Mother with an infant, (b) Pediatrician, Dr. Anjoo Bhatnagar, explaining longitudinal fingerprint study to the mothers while the authors are acquiring an infant's fingerprints. Parents also sign a consent form approved by the Institutional Review Board (IRB) of our organizations.}
    \label{fig:data_collection}
\end{figure}

Fingerprints (Fig.~\ref{fig:frontpage}) are the most promising biometric trait for infant recognition for several reasons. Biological evidence suggests fingerprint patterns are physiologically present on human fingers at birth~\cite{birth1},~\cite{birth2},~\cite{fingerprint_fetus}. While the friction ridge patterns on our fingers may grow or fade over time, the characteristics of the pattern remains unchanged, and longitudinal studies on fingerprint recognition for adults~\cite{yoon} and infants~\cite{infant_jain} show that the fingerprint recognition accuracy does not change appreciably. Additionally, fingerprints are the most convenient, acceptable, and cost-effective biometric to capture from infants (Fig.~\ref{fig:data_collection}).

However, fingerprint recognition of infants comes with its own challenges. First, the fingerprint reader must be very compact (enabling the operator to quickly maneuver the device around the infant), high resolution (due to small inter-ridge spacings), low cost (enabling use in developing countries), ergonomically designed (enabling placement of the infant finger on the platen), and fast capturing (reducing the motion blur). Furthermore, the fingerprint matcher must (i) accomodate heavy non-linear distortions (due to soft infant skin), and (ii) accept high resolution images (1,900 ppi in our case) as input, since infant fingerprints can not be captured with sufficient fidelity at 500 ppi\footnote{The ridge spacing at 500 ppi for adult fingerprint images is about 9-10 pixels compared to 4-5 pixels for infant fingerprint images.}. Current commercial matchers only operate on 500 ppi images since the friction ridge patterns of adults can be easily discriminated at 500 ppi. 

Among various published studies related to infant prints~\cite{galton},~\cite{tno},~\cite{biodev},~\cite{ultrascan},~\cite{jrc},~\cite{jain}, the most extensive study to date has been by Jain~\etal~\cite{jain}, who showed that with a 1,270 ppi\footnote{PPI (pixels per inch) measures the pixel density (resolution) of digital imaging devices.} resolution reader, it is feasible to recognize infants enrolled at the age of 6 months or older. Jain~\etal further showed that if the child's fingerprint is enrolled at the age of 12 months or later, then commercially available 500 ppi fingerprint readers are adequate to capture good quality fingerprints and successfully match the child fingerprints captured a year later. Since immunization for infants commence within 1-3 months of age~\cite{cdc}, in this study, we evaluate the feasibility of fingerprinting and recognizing infants that are below 3 months of age.

\subsection{Custom 1,900 Fingerprint Reader}
\begin{figure}[!t]
    \centering
    \subfloat[]{\includegraphics[height=0.49\linewidth]{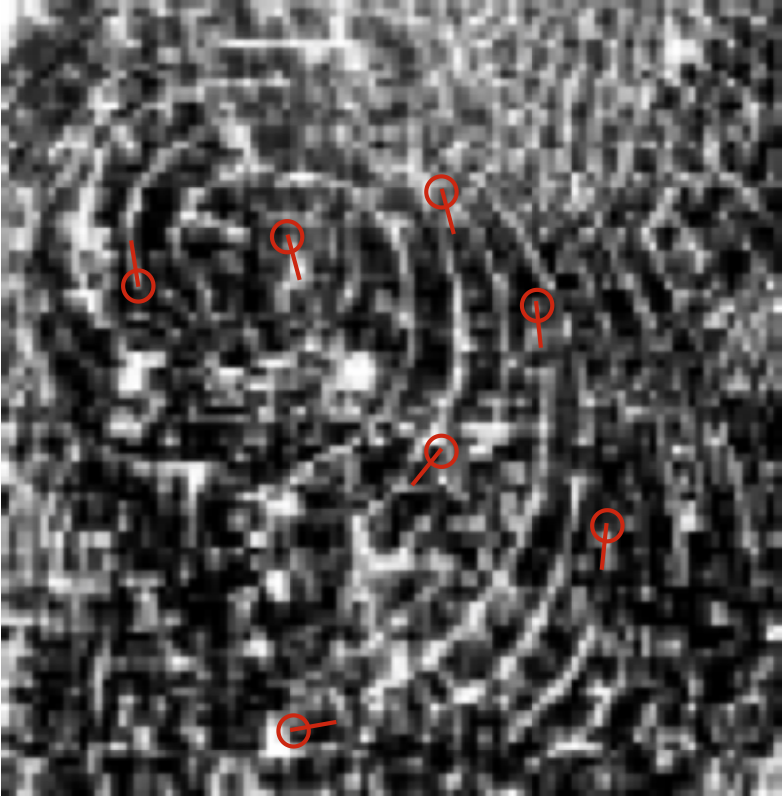}}\hfill
    \subfloat[]{\includegraphics[height=0.49\linewidth]{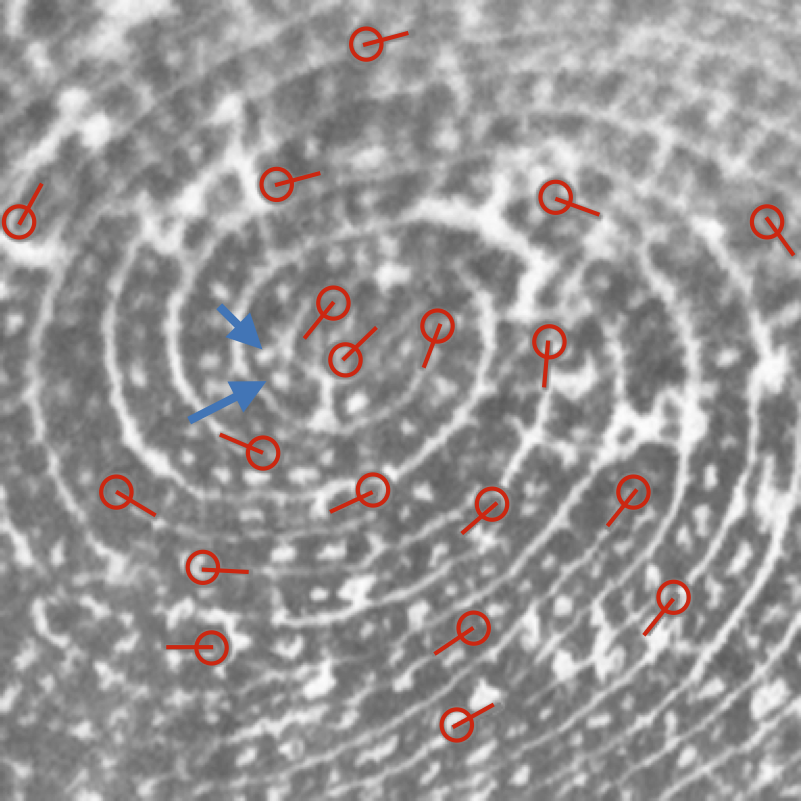}}
    \caption{Effect of fingerprint resolution. (a) Fingerprint of a 2-week old infant captured by a 500 ppi commercial reader; (b) Fingerprint of the same baby by our custom 1,900 ppi, compact, and low cost fingerprint reader. Manually annotated minutiae are shown in red circles (location) with a tail (orientation). Blue arrows denote pores in the infant's 1,900 ppi fingerprint image.}
    \label{fig:resolutions}
\end{figure}

\begin{figure}[!t]
    \centering
    \subfloat[]{\includegraphics[width=0.666\linewidth]{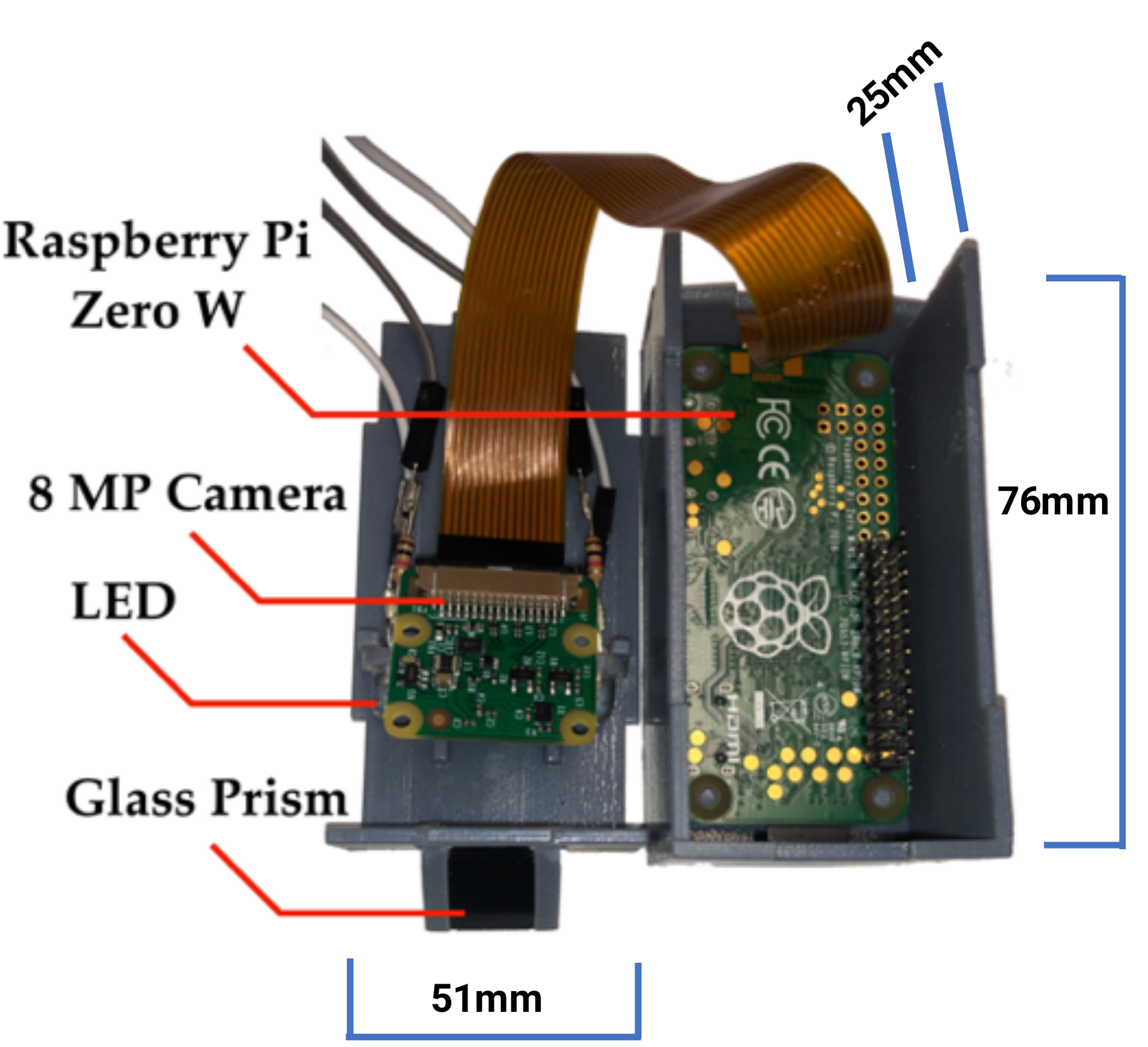}}\hfill
    \subfloat[]{\includegraphics[width=0.333\linewidth]{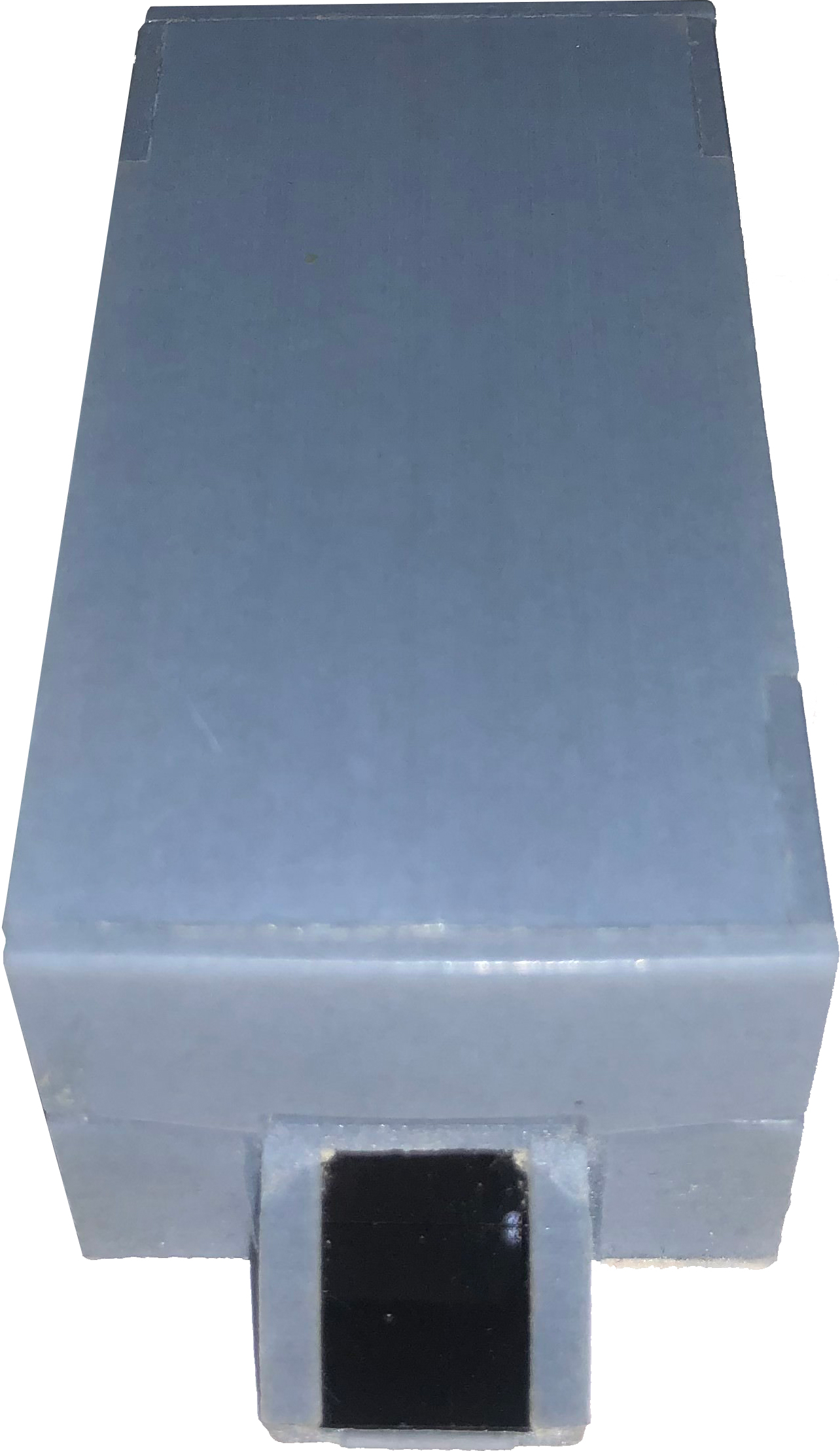}}
    \caption{Prototype of the 1,900 ppi compact (25mm x 51mm x 76mm), ergonomic fingerprint reader. It uses off-the-shelf components (except for the casing), with a total cost of USD 85. During capture, an infant's finger is placed on the glass prism with the operator applying slight pressure on the finger. The fingerprint is transferred to a mobile phone via bluetooth where the fingerprint can be either authenticated or searched against a database (de-duplication). The capture time is 500 milliseconds. The prototype can be assembled in less than 2 hours. See the video at~\url{http://bit.do/RaspiReader}.}
    \label{fig:reader}
\end{figure}

High resolution commercial fingerprint readers, to the best of our knowledge, only reach 1,000 ppi and are incredibly bulky and costly. This motivated us to construct a first-of-a-kind, 1,900 ppi fingerprint reader~(Fig.~\ref{fig:reader}) enabling capture of high-fidelity infant fingerprints (Fig.~\ref{fig:resolutions}). Unlike~\cite{engelsma1},~\cite{engelsma2}, both the size and cost of the reader has been significantly reduced. Furthermore, this fingerprint reader has a glass prism towards the front of the reader (Fig.~\ref{fig:reader}) rather than flush with the top of the reader (as is the case with commercial readers). Since infants frequently clench their fists and have very short fingers, placing the prism out front significantly eases the difficulty of placing an infant's finger on the platen.

In line with our goal of making infant fingerprint recognition ubiquitous and affordable in developing countries, the entire design and 3D parts for the reader casing along with step by step assembly instructions are open sourced.\footnote{\url{https://github.com/engelsjo/RaspiReader}} Figure~\ref{fig:resolutions} shows that this custom 1,900 ppi fingerprint reader is able to capture the minute friction ridge pattern of a 2-week old infant (both minutiae and pores) better than 500 ppi U.are.U. 4500 reader.

\newcolumntype{Y}{>{\centering\arraybackslash}X}
\begin{table}[!t]
\centering
\caption{Infant Longitudinal Fingerprint Dataset Statistics.}
\label{tab:dataset_statistics}
\begin{threeparttable}
\begin{tabularx}{\linewidth}{l|Y}
\noalign{\hrule height 1.5pt}
\textbf{\# Sessions} & 3\\ \hline
\textbf{\# Infants}  & 194\tnote{*}\\ \hline
\textbf{Total \# Images} & 1,724 \\ \hline
\textbf{\# Infants Repeated in Session 3} & 78 \\ \hline
\textbf{Age at Enrollment} & 0 - 3 mos.\\ \hline
\textbf{Time Lapse} & $\sim$3 mos.\\ \hline
\textbf{Male to Female Ratio} & 43\% to 57\%\\ \hline
\noalign{\hrule height 1.5pt}
\end{tabularx}
\begin{tablenotes}\footnotesize
\item[*]Out of the 194 subjects, 118 were present during sessions 1 and 2, and 76 are new infants from session 3 without any longitudinal data. In addition to the three fingerprint acquisition sessions already completed, two additional sessions are planned in September and December, 2019.
\end{tablenotes}
\end{threeparttable}
\end{table}

\begin{figure}[!t]
    \centering
    \includegraphics[width=0.8\linewidth]{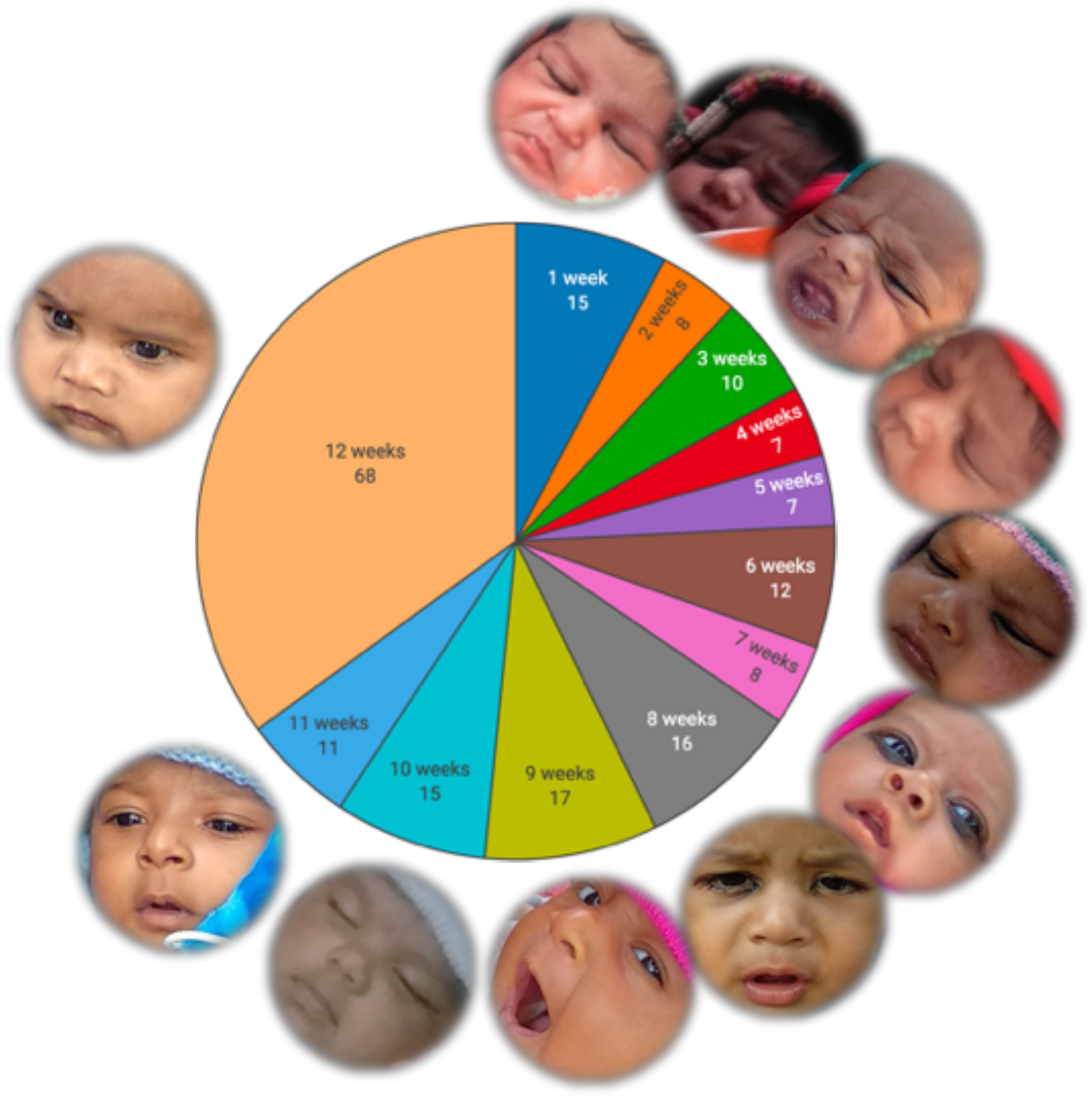}
    \caption{Infant's age at enrollment (1-12 weeks) in our database.}
    \label{fig:age_at_enr}
\end{figure}

\subsection{Infant Longitudinal Fingerprint Dataset}
In order to effectively demonstrate the utility of any infant fingerprint recognition system, we must be able to show its ability to recognize a child based on fingerprints acquired at least a year after the infant's enrollment. That is why collecting a longitudinal fingerprint dataset where the fingerprints of the same child are collected over time is required. 

We collected a dataset comprised of longitudinal fingerprints of 194 infants (0-3 months of age) at Saran Ashram hospital in Dayalbagh, India on December 12-18, 2018 (see Figure~\ref{fig:data_collection})\footnote{The fingerprint dataset cannot be made publicly available per the IRB regulations and parental concerns.}. The infants were patients of the pediatrician, Dr. Anjoo Bhatnagar (Figure~\ref{fig:bhatnagar}). Prior to data collection, the parents were required to sign a consent form (approved by authors' institutional review board and the ethics committee of Saran Ashram hospital).

For 78 infants in the dataset so far, we acquired six impressions from each of the two thumbs (two impressions per thumb per session), over three different sessions. Sessions \#1 and \#2 were separated by 2-3 days and sessions 2 and 3 were separated by about 3 months. Out of the 194 total infants, 118 infants were present in sessions 1 and 2, and 78 of them came back for session 3. In session 3, 76 new infants were enrolled whose fingerprints were used for training data. During collection, a dry or wet wipe was used, as needed, to clean the infant's finger prior to fingerprint acquisition. On average, data capture time, for 4 fingerprint images (2 per thumb) and a face image per infant, was 3 minutes. This enabled high throughput during the in-situ evaluation, akin to the operational scenario in immunization and nutrition centers.  

Longitudinal fingerprint dataset statistics are given in Table~\ref{tab:dataset_statistics}. Figure~\ref{fig:age_at_enr} shows the age distribution of the infants for the three sessions undertaken so far (December 12-18, 2018 and March 3-9, 2019).

\begin{figure*}[!t]
    \centering
    \includegraphics[width=\linewidth]{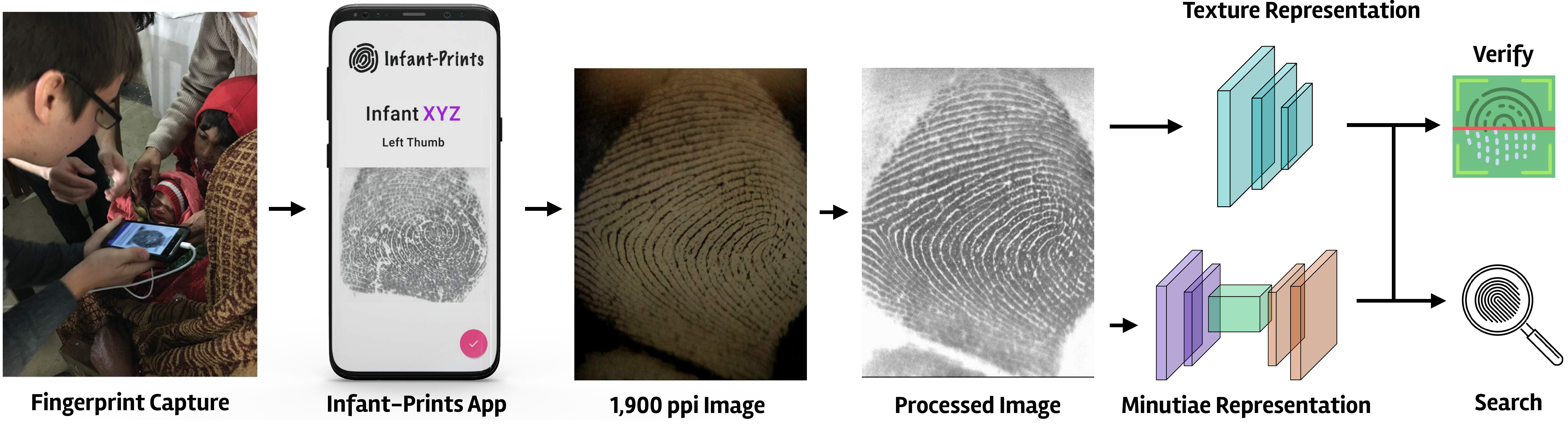}
    \caption{Overview of the Infant-Prints system.}
    \label{fig:overview}
\end{figure*}

\section{Infant Fingerprint Matching}

State-of-the-art fingerprint feature extractors and matchers are designed to operate on 500 ppi adult fingerprint images. This limitation forced the authors in~\cite{jain} to down-sample the fingerprint images captured at 1,270 ppi to enable compatibility with COTS (Commercial Off The Shelf) matchers. In this study, we develop a custom Convolutional Neural Network based matcher which directly operates on 1,900 ppi fingerprint images so that we do not have to down-sample images and discard valuable discriminative cues available in high resolution images. 

The fingerprint comparison score is based on the fusion of (i) CNN-based custom texture matcher and (ii) two state-of-the-art COTS matchers. Note that we do need to downsample the 1900 ppi images for the two COTS matchers as was done in~\cite{jain}.

\subsection{Texture Matcher}

Engelsma~\etal~\cite{engelsma3} proposed a CNN architecture embedded with fingerprint domain knowledge for extracting discriminative fixed-length fingerprint representations. Inspired by the success of the network to learn additional textural cues that go beyond just minutiae points, we adopt this matcher for infant fingerprint recognition. In particular, we modify the network architecture as follows: (i) the input size of 448 x 448 is increased to 1024 x 1024 (through the addition of convolutional layers) to support 1,900 ppi images and (ii) the parameters of the added convolutional layers and the last fully connected layer are re-trained on the 1,270 ppi (upsampled to 1900 ppi) longitudinal infant fingerprints acquired by Jain~\etal in~\cite{jain} combined with 500 of our 1900 ppi images which we set aside for training. In total, we re-train the network with 9,683 infant fingerprint images from 1,814 different thumbs. 

During the authentication or search stage, the CNN accepts a 1,900 ppi infant fingerprint as input and outputs a 512-dimensional fixed-length representation of the fingerprint. This representation can be compared to previously enrolled representations via the cosine distance between two given representations at 600K comparisons/second on a commodity Intel i5 processor with 8 GB of RAM. 

\subsection{COTS matchers}

We fuse two state-of-the-art COTS matchers (COTS-A and COTS-B\footnote{COTS-B matcher is one of top-three performers in the
NIST ELFT-EFS evaluations~\cite{cots_1},~\cite{cots_2}. Due to NDA, we cannot disclose the vendors' names.}). COTS-B is specifically designed for latent fingerprints (whose properties are similar to infant fingerprints in terms of small ridge area and image distortion), while COTS-A is designed for plain (slap) prints.

\section{Android Application}

To make Infant-Prints portable and operator friendly, we develop an Android Application. The Android App (i) receives 1,900 ppi images from the fingerprint reader over bluetooth, and (ii) performs fingerprint verification (1:1 comparison) or identification (1:N search). After a successful match or search, subject's meta-data such as vaccination records can be displayed on the mobile phone.

\section{Experimental Protocol}

\newcommand{\specialcell}[2][c]{%
  \begin{tabular}[#1]{@{}c@{}}#2\end{tabular}}
  \newcommand{\tabitem}{~~\llap{\textbullet}~~}

\begin{table*}[!t] 
\footnotesize
\caption{Verification accuracy (TAR (\%) @ 0.1\% and 1.0\% FAR) for infants older than (a) 0 months, (b) 1 months, and (c) 2 months. Fingerprint impressions from session 1 and sessions 2 are compared to session 3.}
\centering
\begin{tabular}{c||c|c||c|c||c|c}
\noalign{\hrule height 1.5pt}
\textbf{Age at Enrollment} &\multicolumn{2}{c||}{\textbf{\specialcell{Matcher: COTS \\ Images: 500 ppi}}} & \multicolumn{2}{c||}{\textbf{\specialcell{Matcher: COTS \\ Images: 1,900 ppi}}} & \multicolumn{2}{c}{\textbf{\specialcell{Matcher: COTS + Texture \\ Images: 1,900 ppi}}}\\ \noalign{\hrule height 1.2pt}
& \textbf{0.1\% FAR} & \textbf{1.0\% FAR} & \textbf{0.1\% FAR} & \textbf{1.0\% FAR} &\textbf{0.1\% FAR} & \textbf{1.0\% FAR}\\ \hline
0 - 3 months (78 infants) & 57.5 & 65.8 & 64.1 & 67.9 & 66.7 & 78.2 \\
1 - 3 months (69 infants) & 64.0 & 70.3 & 68.1 & 73.9 & 75.4 & 85.1\\
2 - 3 months (51 infants) & 74.5 & 78.7 & 82.4 & 86.3 & 90.2 & 94.1\\
  \hline
\noalign{\hrule height 1.5pt}
\end{tabular}
\label{tab:verification_age}
\end{table*}

Our experiments are designed to show the benefits of (i) our 1,900 ppi reader over the baseline 500 ppi readers, (ii) a high-resolution texture-based matcher for 1,900 ppi images, and (iii) the feasibility of recognizing infants under the age of 3 months.

In all of our matching experiments, we fuse the scores from the two thumbs of an infant. Additionally, we fuse the scores from multiple impressions of the same thumb. In particular, the enrollment template of a thumb is comprised of 2-4 impressions (depending on infant's cooperation during capture), captured during sessions 1 and 2 in December, 2018. The probe template for a given thumb is comprised of two new impressions captured in March, 2019. Finally, the multiple impressions comprising the enrollment and probe templates are compared and the scores fused into one final score. For both thumb level fusion and impression level fusion, we simply average the scores. 

Given the high throughput of our system, collecting multiple fingerprints from both thumbs is easily accomplished in an operational testing scenario. Since children are frequently placing their thumbs in the mouth (causing wet fingers) and moving their hands during collection (causing motion blur), impression-level fusion from both the thumbs is necessary to ensure accurate and reliable recognition. 

\section{Performance Comparison}

\begin{figure}[!t]
    \centering
    \subfloat[1 month old]{\includegraphics[height=1.95in]{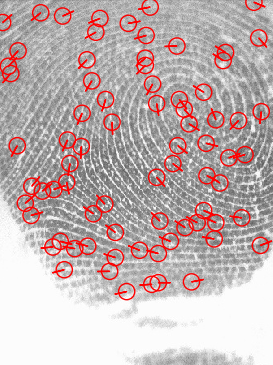}}\hfill
    \subfloat[3 month old]{\includegraphics[height=1.95in]{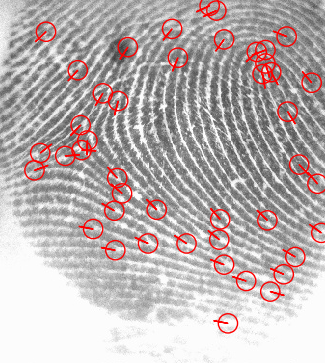}}
    \caption{A False Match due to high inter-fingerprint similarity likely due to small ovelapping area and non-linear distortion which has changed the ridge spacing. The minutiae locations and orientations have been annotated by COTS-A.}
    \label{fig:fm}
\end{figure}

\begin{figure}[!t]
    \centering
    \subfloat[]{\includegraphics[height=1.95in]{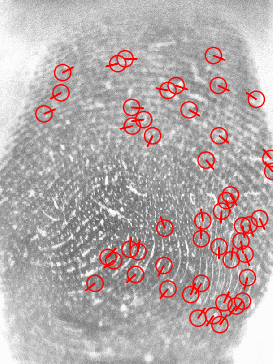}}\hfill
    \subfloat[]{\includegraphics[height=1.95in]{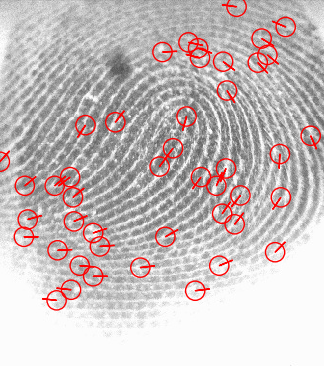}}
    \caption{A False Non-Match due to the growth of the fingerprint size in (a) over a 3-month period to the fingerprint in (b). Additionally, the fingerprint in (a) contains missing and spurious minutiae due to the low quality of the 2 month old infant's fingerprints. The minutiae locations and orientations have been annotated by COTS-A.}
    \label{fig:fnm}
\end{figure}

The performance of Infant-Prints is reported as an ablation study in Table~\ref{tab:verification_age}. From these results, we make several observations. First, we note that the 1,900 ppi fingerprint images boost the recognition performance even when using low-resolution COTS matchers (operating on 500 ppi images, i.e. the 1900 ppi images first had to be down-sampled). Next, we show that by fusing our high resolution matcher (1900 ppi) with existing COTS matchers we are able to significantly boost the recognition accuracy. Examples of False Matches and False Non Matches are shown in Figures~\ref{fig:fm} and~\ref{fig:fnm}.

Most importantly we show, for the first time, that it is feasible to recognize children, under the age of 3 months, by their fingerprints, over a time lapse of 3 months. When the minimum age at enrollment is set to 1-month, we obtain a TAR of 85.1\% @ FAR = 1.0\%. When bumping the minimum age of enrollment up to 2-months, the recognition accuracy improves to 94.1\% @ FAR = 1.0\% (see Table~\ref{tab:verification_age}). With enrollments at 1-2 months, children are tied to a longitudinal identity just in time for first vaccinations and proper nutritional supplements. Therefore, it is our hope that Infant-Prints, after further accuracy improvements, can be used to significantly alleviate child suffering and death around the world.  

\section{Conclusion}

We have introduced a complete infant fingerprint recognition system, called Infant-Prints, comprised of a custom, 1,900 ppi fingerprint reader, a new texture-based infant fingerprint matching algorithm, and an Android application for operator use for viewing infant's meta data in real-time. In a longitudinal in-situ evaluation, we show that Infant-Prints is capable of reliably and accurately identifying infants under the age of 3 months (TAR of 94.1\% @ FAR = 1.0\% above the age of 8 weeks). We have shown that the same low-cost, portable and high resolution Infant-Prints prototype can also be used to identify adults with high accuracy~\cite{engelsma2}. Our ongoing study is addressing (i) improvements in fingerprint reader design and capture speed, and (ii) improving the accuracy and robustness of fingerprint matcher. Our goal is to transfer the Infant-Prints prototype system to an organization for larger in-situ evaluation and deployment. 

{\small
\bibliographystyle{unsrt}
\bibliography{egbib}
}

\end{document}